\documentclass[10pt,twocolumn,letterpaper]{article}

\usepackage{iccv}
\usepackage{times}
\usepackage{epsfig}
\usepackage{graphicx}
\usepackage{amsmath}
\usepackage{amssymb}
\usepackage{multirow}
\usepackage{balance}

\usepackage[breaklinks=true,bookmarks=false]{hyperref}

\iccvfinalcopy 

\def\iccvPaperID{****} 

\ificcvfinal\fi

\begin{document}

\title{CLIP4Caption ++: Multi-CLIP for Video Caption}

\author{Mingkang Tang$^{1,2}$, Zhanyu Wang$^{1}$, Zhaoyang Zeng$^1$, Fengyun Rao$^1$, Dian Li$^1$\\
$^1$Kandian Content AI Lab, Platform and Content Group, Tencent\\
$^2$Tsinghua Shenzhen International Graduate School, Tsinghua University\\
{\tt\small {{mktang,zhanyuwang}@tencent.com}}
}

\maketitle
\ificcvfinal\thispagestyle{empty}\fi
\begin{abstract}
This report describes our solution to the VALUE Challenge 2021 in the captioning task. Our solution, named CLIP4Caption++, is built on X-Linear/X-Transformer, which is an advanced model with encoder-decoder architecture. We make the following improvements on the proposed CLIP4Caption++:
1) we utilize three strong pre-trained CLIP models to extract the text-related appearance visual features. 2) we adopt the TSN sampling strategy for data enhancement. 3) we involve the video subtitle information to provide richer semantic information.
4) we design word-level and sentence-level ensemble strategies. Our proposed method achieves 86.5, 148.4, 64.5 CIDEr scores on VATEX, YC2C, and TVC datasets, respectively, which shows the superior performance of our proposed CLIP4Caption++ on all three datasets.

\end{abstract}
\section{Introduction}

Video captioning is an advanced multi-modal task that automatically generates a natural language sentence of a given video clip. Currently, numerous SOTA approaches for multi-modal tasks (such as image-text retrieval, image captioning, etc.) adopt the training paradigm of "pre-training + fine-tune." Common pre-training tasks include mask language model (MLM), mask region model(MRM), video-text matching (VTM), etc., which have been proved very helpful to improve downstream tasks. The essence of those pre-training tasks is to align the two modalities in the metric space. From this point of view, in feature selection, we utilize multiple strong CLIP pre-training models to extract video appearance features because CLIP's pre-training model has the natural ability to align image and text in its feature space since it has trained on hundreds of millions of image-text pairs. Specifically, we use three CLIP pre-training models for video appearance features, SlowFast pre-training model for video motion features, and discard the S3D and Resnet features that are officially provided. We did not use the pre-training method or add additional data in our whole training process. We also made several effective improvement measures: data enhancement, introducing subtitle information to enhance the feature, an ensemble strategy designed especially for the generation tasks. The proposed CLIP4Caption++ solution won the championship on the VALUE~\cite{li2021value} competition captioning task. 

 
  
 
 

\section{Methods}

\begin{figure*}
  \centering
  \includegraphics[width=\textwidth]{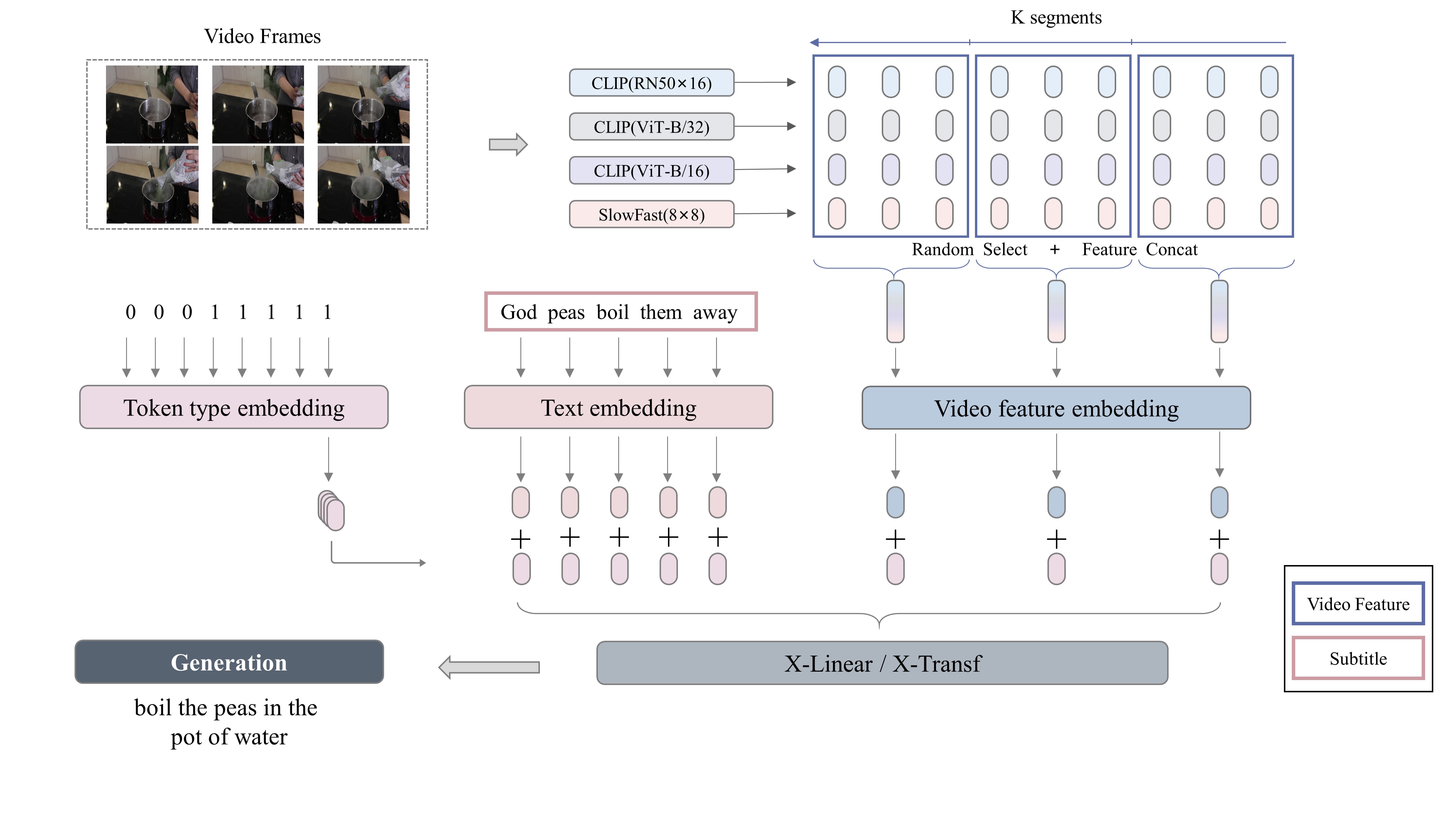}
  \caption{An Overview of our proposed CLIP4Caption++ framework.}
  \label{fig1}
\end{figure*}
Figure\ref{fig1} illustrates the framework of our proposed CLIP4Caption++ for video captioning. First, we use the pre-trained model of CLIP and SlowFast to extract visual features from origin video (\ref{section1}) (higher part in Fig.~\ref{fig1}). Second, we sampled the extracted frame features using TSN~\cite{wang2016temporal} sampling (\ref{section2}). And then, video feature and text, as well as the token type, were embedded by embedding layer for bimodal feature fusion (\ref{section3}) (lower part in Fig.~\ref{fig1}). For ensemble, we train multiple caption models with the different settings and ensemble the output of multiple models for a final strong result (\ref{section4}). Details will be elaborated as follows.
\subsection{Feature Extraction}\label{section1}
We use the pre-trained CLIP and SlowFast as our video feature extractor. We extract CLIP-based video features at 10 FPS. CLIP~\cite{radford2021learning} was pre-trained on image-text matching with 400 million image-text pairs, making it vital to learn image representations. Furthermore, CLIP has two different architectures for the image encoder. For Resnet~\cite{he2016deep} architecture, we chose 16x the compute of a ResNet-50, denoted as CLIP(RN50x16); For the Vision Transformers~\cite{dosovitskiy2020image} architecture, we chose ViT-B/32 and ViT-B/16 as the image encoder, denoted as CLIP(ViT-B/32) and CLIP(ViT-B/16) separately.

SlowFast~\cite{feichtenhofer2019slowfast} was well designed for video recognition, capturing spatial semantics and motion in different frame rates. We chose SlowFast 8×8, ResNet-50 as our feature extractor; the Slow pathway and Fast pathway take 8 and 32 frames as the network input separately, both sampled from a 64-frame raw clip. We sampled each 64-frame raw clip in a sliding time window. The length of the time window is 1.5 seconds, and the move stride is 0.3 seconds. The sliding time window allows more clips so that SlowFast can sense a longer time range of video frames without reducing the total number of extracted features.

\subsection{TSN Sampling Strategy}\label{section2}
After feature extraction, we sample the extracted frames according to TSN~\cite{wang2016temporal}. During training, TSN sampling divides all frames into K segments of equation duration and then randomly samples one frame from each segment, thus increasing the sample randomness on the limited frame length. After using TSN sampling, the frames selected in each epoch are inconsistent, accordingly achieving data enhancement for video data. 
The TSN sampling is not used in testing, as we evenly extract K frames from all frame features at equal intervals.

\begin{table*}[t]
\centering
\setlength{\tabcolsep}{6mm}{
\begin{tabular}{ccccccc}
\hline
\multirow{2}{*}{Datasets} & \multirow{2}{*}{Model} & \multicolumn{3}{c}{Public Test} &  & Private Test \\ \cline{3-5} \cline{7-7} 
                          &                          & BLEU-4   & ROUGE-L   & CIDEr    &  & CIDEr        \\ \hline
\multirow{2}{*}{TVC~\cite{lei2020tvr}}      & single model             & 13.43    & 35.73     & 61.20     &  & 59.41        \\
                          & ensemble                 & 15.03    & 36.86     & 66.02    &  & 64.49        \\ \hline
\multirow{2}{*}{YC2C~\cite{zhou2018towards}}     & single model             & 13.03    & 39.78     & 136.01   &  & 133.98       \\
                          & ensemble                 & 13.60    & 41.43     & 148.91   &  & 148.39       \\ \hline
\multirow{2}{*}{VATEX~\cite{wang2019vatex}}    & single model             & 38.66    & 53.61     & 81.95    &  & 83.67        \\
                          & ensemble                 & 40.59    & 54.45     & 85.71    &  & 86.53        \\ \hline
\end{tabular}}
\vspace{1mm}
\caption{Best single model and multiple model ensemble results on the TVC, YC2C and VATEX test set of public and private.}
\label{table2}
\end{table*}
\subsection{Bimodal Feature Fusion}\label{section3}
Frames sampling makes the four different features to the same frame length so that we can concatenate them to the final 4096-dim visual video feature. As shown in the Fig.~\ref{fig1}, the visual video feature was embedded by the video embedding layer. We find that the video subtitle can provide rich semantic information, which can benefit the caption generation progress. Thus we also leverage the video subtitle to enhance our models. The subtitle sentences are tokenized and embedded by the text embedding layer. Finally, the text embedding and video embedding are concatenated as the X-Linear/X-Transformer input. For making the model distinguish between these two embeddings, a particular embedding layer, token type embedding layer, was added. First, we set the video label as 0 and the text label as 1. And then, the token type embedding layer is used to embedded video labels and text labels. After that, token type embeddings are added to the concatenated video embedding and text embedding. 

\begin{table}[h]
\setlength{\tabcolsep}{1mm}{
\begin{tabular}{lcc}
\hline
Method                                  & \multicolumn{1}{l}{CIDEr} & \multicolumn{1}{l}{gain(\%)} \\ \hline
C32+SF                                  & 65.15                     & -                            \\
C32+SF+ST                               & 65.67                     & 0.8                         \\
C32+C16+C50+SF+ST                       & 69.03                     & 6.0                          \\
C32+C16+C50+SF+ST+TSN                   & 69.23                     & 6.3                          \\
C32+C16+C50+SF+ST+TSN+SCST                & 80.82                     & 24.1                         \\
C32+C16+C50+SF+ST+TSN+SCST+ES             & 85.71                     & 31.2                          \\ \hline
\end{tabular}}
\vspace{1mm}
\caption{Performance gain from each component. ``C32", ``C16", ``C50", ``SF", ``ST", ``TSN" and ``ES" stand for ``CLIP(Vit-B/32)", ``CLIP(Vit-B/16)", ``CLIP(RN50x16)", ``SlowFast(8x8)", ``Subtitle", ``TSN sampling strategy" and ``ensemble", respectively.}
\label{table1}
\end{table}
\subsection{Ensemble Strategy}\label{section4}

In order to achieve a more powerful caption result, we design a word-level ensemble strategy and a sentence-level strategy for the captioning task. For word-level strategy, we ensemble the output of all the models in each generation step. During the generation, the captioning model generates one word according to the softmax output and takes the word as the input for the next step. We averaged the softmax output of all models and chose the word with max probability in each step. The model selects the correct word with a greater probability after considering all the results of multiple models, reducing the impact of incorrect words on subsequent generated results.

For sentence-level strategy, we utilize the CIDEr metric as the ``importance score'' of a generated sentence and select the highest score to compose the final result. Mathematically, taking the predicted captions of one video from $n$ different models as $T_{i}$, the importance score for $ith$ caption $S_{i}$ using CIDEr metric can be calculated by assuming the rest of the predicted captions as ``ground-truth" captions:
    \begin{equation*}
    \label{eq:1}
    S_{i} = \mathbf{CIDEr}(ref=\left[\mathbf{T}/\mathbf{T}_i\right], hpy=\mathbf{T}_{i}),
    \end{equation*}
    where $i \in \left[1, n\right]$ and $\mathbf{CIDEr}(\cdot)$ is the function for computing CIDEr score, ``$\mathbf{T}/\mathbf{T}_i$" means exclude $\mathbf{T}_i$ from the set $\mathbf{T}$. The predicted caption with largest score $S$ is selected as the final output. 

We use the word-level strategy first for a better single caption result and add the word-level ensemble result to the other single-model results for the sentence-level ensemble.

\section{Experiments}

\subsection{Implementation Details}
We follow \cite{pan2020x} to conduct a two-stage training for CLIP4Caption++. For the first stage, we train the primitive CLIP4Caption++ with our enhanced video embedding. We employ the X-Linear/X-Transformer attention block to encode enhanced video embedding sequences and further used it in the decoder to mine the interaction information between different modes. Training stage one is done on 4 NVIDIA Tesla P40 GPU graphics, with batch size set to 128 and max epochs set to 100.

For the second stage, we train the caption model with SCST~\cite{rennie2017self}. SCST utilized the policy-gradient method, which takes the caption Cider in the training stage as a reward score for caption generation, and uses the current model's reward in the test stage as the reward baseline. Thus, SCST optimizes CIDEr directly and encourages the model to perform consistently in training and testing stages. We initialize the model with the weights from the best model in stage one and set batch size as 20 and total epochs as 60, with a single GPU. Experimental result shows that SCST significantly improved the CIDEr score in stage two. 


\subsection{Ablation Study}
We report the performance gain brought by each component. All the results were based on X-Linear. As shown in the Table~\ref{table1}, Each strategy contributed to the CIDEr improvement, and the concatenated visual video feature and SCST(Training stage two) provided a significant improvement. The enhanced video feature allows the model to capture more video semantic information, and SCST makes the model directly optimize the CIDEr metric.

\subsection{Result}

We train multiple models with different frame lengths, combinations of features, and other hyperparameters. We adopt the standard captioning metrics including BLEU-4~\cite{papineni2002bleu}, ROUGE-L~\cite{lin2004rouge}, CIDEr~\cite{vedantam2015cider} to evaluate the performance of the captioning models. We report the single model result and ensemble result on the public test set and validation test set of TVC~\cite{lei2020tvr}, YC2C~\cite{zhou2018towards}, VATEX~\cite{wang2019vatex} datasets in Table\ref{table2}. From Table\ref{table2} we can find that our ensemble strategy brings significant improvement on all datasets, and the final result ranks 1st in the VALUE competition.

\section{Conclusion}
This work focuses on better video embedding and proposes the CLIP4Caption++, which improves video embedding by enhanced visual video features and multi-modal text fusion. Besides, we introduce word-level and sentence-level ensemble strategies to ensemble multiple models' captioning results. Extensive experiments indicate that each strategy contributed to the CIDEr improvement. Our method rank first in the VALUE competition in captioning task.

{\small
\bibliographystyle{ieee_fullname}
\balance
\bibliography{egpaper_final}
}

\end{document}